\documentclass[a4paper, 10pt, conference]{ieeeconf} 
\usepackage{FG2020}
\FGfinalcopy 

\IEEEoverridecommandlockouts                
\usepackage[noadjust]{cite}
\usepackage[nolist]{acronym}                    
\usepackage{algorithm}
\usepackage{algorithmicx}
\usepackage{algpseudocode}
\usepackage{pifont}
\usepackage{cancel}
\usepackage{amsfonts}       
\usepackage{amsmath}        
\usepackage{amssymb}        
\usepackage{mathtools}
\usepackage{booktabs}       
\usepackage{multirow}
\usepackage{graphicx}
\usepackage[utf8]{inputenc}
\usepackage{siunitx}		
\usepackage{subcaption}     
\usepackage{tabularx}                   
\usepackage{url,xspace,boxedminipage}   
\usepackage{placeins}
\usepackage{pdflscape}
\usepackage{balance}
\usepackage{hyperref}
\usepackage{setspace}
\usepackage{tikz}
\usetikzlibrary{calc}

\newcommand{\positiontextbox}[4][]{%
  \begin{tikzpicture}[remember picture,overlay]
    \node[inner sep=5pt,right, fill=white,draw,line width=1pt,#1] at ($(current page.north west) + (#2,-#3)$) {#4};
  \end{tikzpicture}%
}

\def\FGPaperID{} 

\title{
\LARGE \bf
Continual Learning for Affective Computing}
\author{\parbox{16cm}{\centering
    {Nikhil Churamani}\\
    {
    Department of Computer Science and Technology, \\University of Cambridge, United Kingdom\\
    \tt{\small nikhil.churamani@cl.cam.ac.uk}}}
    \thanks{N. Churamani is supported by the EPSRC under the Cambridge EPSRC-DTP EP/R513180/1 project ref. 2107412.}
}

\begin{document}

\pagestyle{plain}
\maketitle

\begin{abstract}

Real-world application requires affect perception models to be sensitive to individual differences in expression. As each user is different and expresses differently, these models need to personalise towards each individual to adequately capture their expressions and thus, model their affective state. Despite high performance on benchmarks, current approaches fall short in such adaptation. In this work, we propose the use of \textit{\acf{CL}} for affective computing as a paradigm for developing personalised affect perception.
    
\end{abstract}

\positiontextbox[fill=white]{0.65cm}{1cm}{Accepted at the Doctoral Consortium for the IEEE International Conference on Automatic Face and Gesture Recognition (FG), 2020.}

\section{Introduction}
Current approaches in affect perception predominantly focus on instantaneous (frame-based) analysis of human behaviour. They rely on glimpses of heightened audio-visual stimuli to infer the affective state of 
users~\cite{Sariyanidi2017Learning, PORIA2015104}. Even though this works well in providing a short-term evaluation of human expression, where only a snapshot of user behaviour is required, analysing long-term interactions, under varying affective contexts, is still an open problem~\cite{Sariyanidi2015Automatic}. As a result, despite current (deep) learning approaches achieving high performance scores on expression recognition benchmarks (see~\cite{Sariyanidi2017Learning, nott44740, PORIA201798} for an overview), they are not able to sufficiently model the dynamics of human affective behaviour during long-term interactions.

The development cycle for most (deep) learning approaches follows a fixed transition from first being trained in isolation on a `large enough' dataset with high variability, and then being applied to real-world applications~\cite{Schmidhuber15}. With a lot of the existing datasets capturing \textit{posed} expressions recorded in fixed laboratory conditions, generalisation to real-world scenarios becomes problematic~\cite{PORIA201798} as the real-world may be very different to the conditions under which these datasets are recorded. As a result, the research has turned towards training and testing models on data that captures affect \textit{in-the-wild}~\cite{Dhall2014}, containing samples collected from real-world scenarios. Yet, these models still follow the same development cycle, with little to no adaptability in their application, facing difficulties in capturing individual nuances in human affective expression. 

Thus, there is a need for models to adapt to individual differences in expression, enabling them to personalise towards individuals, in real time. \textit{Personalisation}, in this context, can be understood as the \textit{ability to account for individual differences in expression, as well as individual behaviour patterns} while sensing and analysing their affective state during an interaction. Despite some efforts focussing on such personalisation to realise generic-to-specific perceptual adaptations~\cite{Rudoviceaao6760, Chu2017Selective}, more work is needed on personalised affect perception. 

\acf{CL} research~\cite{HASSABIS2017245,Parisi2018b} aims to address this very problem of long-term adaptability in agents, enabling them to learn incrementally as they interact with their environment. \ac{CL} models are commonly applied to learning different objects and tasks in an incremental manner~\cite{Parisi2018b}. The basic principles of \ac{CL}, however, can also help in developing models for affect perception~\cite{Lu2019,pmlrbarros19a} that learn to personalise towards different users. This can be particularly beneficial in real-world interactions where social agents, embedded with such affect perception mechanisms, learn and adapt with each user they interact with. Starting from a limited general understanding, they can learn to personalise towards each user, while at the same time, learning global and generic features.

In this work, we propose \textit{Continual Learning} as a learning paradigm for \textit{Affective Computing}. In this paper, in particular, we discuss learning mechanisms that model generic-to-specific adaptations in \ac{FER} models to enhance their personalisation capabilities. Our focus on \ac{CL} approaches for lifelong learning of affect presents a two-fold problem. On the one side, the model should learn to personalise towards a particular user, learning how they express their affective state, yet, at the same time, it also needs to adapt with different users. Thus, learning happens at two-levels, \textit{individual}, that is, learning different expressions of a particular user, and \textit{between-individuals}, that is extending the learning to be sensitive towards different users. In our current work~\cite{Churamani2020CLIFER}, we examine the former, learning different \textit{facial expression} categories for the same individual. Future work will focus on extending this to \textit{between-individual} adaptation, where the same model can be applied to learning with different users.

\section{Work Summary}
\subsection{Proposed Framework}
Our recent work~\cite{Churamani2020CLIFER} presented a novel framework that investigates a \ac{CLS}-based~\cite{Parisi2018a}, neuro-inspired approach for learning facial expressions. The proposed framework for \ac{CLIFER} (see Fig.~\ref{fig:model_complete}) consists of two components: (i)~a generative auto-encoder model for \textit{imagination}, that is, simulating additional facial images for individual subjects for seen and unseen classes to augment learning; and (ii)~a \ac{CLS}-based dual-memory learning model for \ac{FER} that adapts to novel data as well as balances long-term retention of knowledge. The imagination model learns to generate facial images for $6$ expression classes, namely, \textit{anger, happy, fear, sad, surprise} and \textit{neutral}. These generated images augment learning in the dual-memory model that learns to classify these images. The two components of the framework are briefly described here:

\subsubsection{Auto-Encoder-based Imagination Model}

The recent success of generative models~\cite{Han2019Adversarial}, has enabled the simulation of photo-realistic images containing human faces with different emotion expressions~\cite{ding2017exprgan}. Such models can be used to generate additional images for a user by translating images from seen expression classes to generate samples for yet unseen classes. Thus, having seen only a few images of a subject, the model can generate additional data (akin to \textit{imagination} in humans) for the individual to preempt future interactions. For an agent, such a model can be applied to realise \textit{imagined contact}~\cite{Wullenkord2019} with a participant simulating \textit{imagination} as a substitute for sensory experience.

\begin{figure}
    \centering
    \includegraphics[width=0.5\textwidth]{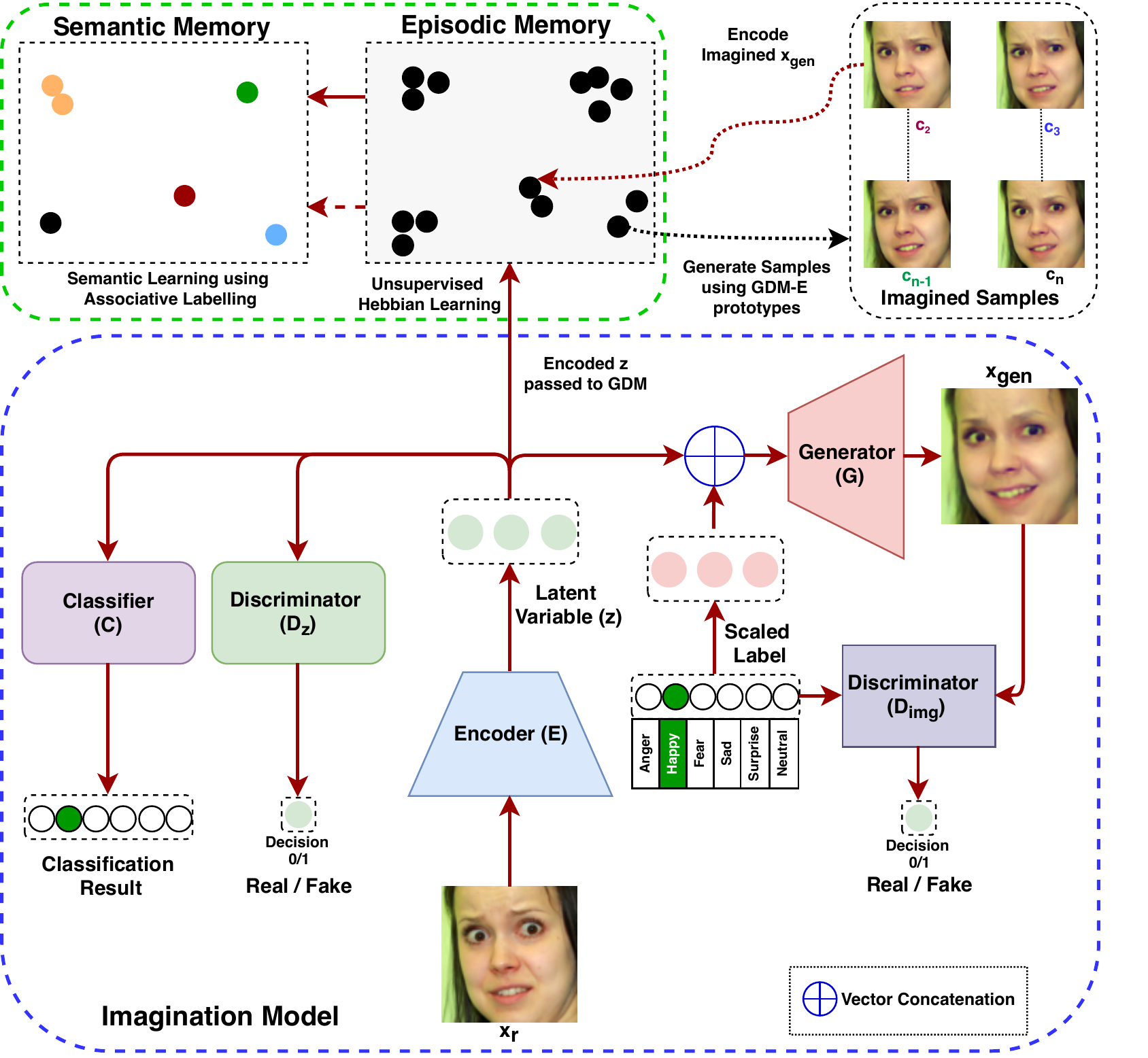}
    \caption{The CLIFER Framework for FER with Imagination: $x_r$ is encoded and passed on to the different models for further processing. Dual-memory model is also trained on the encoded $x_{gen}$.~\cite{Churamani2020CLIFER}}
    \label{fig:model_complete}
\end{figure}

To embed such capabilities in the \ac{CLIFER} framework, we employ a \acf{CAAE}-based~\cite{ding2017exprgan} \textit{imagination} model (see Fig.~\ref{fig:model_complete}) that takes an original (input) image ($x_r$) and generates translated images ($x_{gen}$) for each of the $6$ expressions. 


\subsubsection{CLS-based Dual-Memory Model}
The \ac{GDM} architecture~\cite{Parisi2018a} is used as the basis for incrementally acquiring and integrating knowledge in the \ac{CLIFER} framework. It consists of two hierarchically arranged recurrent \acf{GWR} neural networks~\cite{MARSLAND20021041} representing the \textit{episodic} (\ac{GDM}-E) and \textit{semantic} (\ac{GDM}-S) memories, respectively. 

\begin{itemize}
    \item \textbf{Episodic Memory:} Each input image is encoded by the Encoder (E) and sequentially passed to the \ac{GDM}-E which rapidly learns (using a high learning-rate) non-overlapping representations. This is achieved using a distance-based similarity measure, implementing unsupervised Hebbian-based learning. As it receives data, one class at a time, it creates feature prototypes for each input sample, rapidly adapting to novel data.
    
    \item \textbf{Semantic Memory:} The \ac{GDM}-S learns compact overlapping representations that generalise across samples from a particular class. After each episode (mini-batch) of sequential input, \ac{GDM}-S receives the \textit{winner} neurons from \ac{GDM}-E, along with label annotations. A frequency-based associative labelling scheme~\cite{Parisi2018a} is used to associate feature prototypes with their respective labels with new neurons added to the \ac{GDM}-S only if the existing neurons cannot correctly classify the input. Periodically, neural activation trajectories from \ac{GDM}-E are replayed to both \ac{GDM}-E and \ac{GDM}-S for \textit{pseudo-rehearsal} of past knowledge, mitigating forgetting.
    
    \item \textbf{Imagination:} After receiving data samples from a particular class, \textit{winner} neurons (feature prototypes) from the \ac{GDM}-E are passed to the \textit{imagination} model which simulates facial images of a subject for each expression class, preserving the identity of the subject. These imagined images are encoded and replayed to both \ac{GDM}-E and \ac{GDM}-S, augmenting learning in \ac{CLIFER}.

\end{itemize}

\subsection{Experimentation and Results}
\label{sec:experiments}

We conduct two experiments to evaluate the \ac{CLIFER} framework on its ability to (1)~remember previously seen expression classes for an individual and (2)~to extend its learning to yet unseen facial expressions. \ac{CLIFER} is trained and tested separately for each subject from the \acs{RAVDESS}~\cite{livingstone2012ravdess}, MMI~\cite{Valstar2010idhas} and BAUM-1~\cite{ZhalehpourBaum2017} datasets. While \acs{RAVDESS} and MMI datasets provide an evaluation on \textit{posed} samples, BAUM-1 evaluates the model on \textit{spontaneous} \ac{FER}.

In our experiments we compare four different models, namely; (i)~the \ac{GDM} model without replay, (ii)~\ac{GDM} model with the pseudo-replay mechanism (see~\cite{Parisi2018a} for details), (iii)~the proposed \ac{CLIFER} framework, and (iv)~a \ac{MLP}-based classifier that acts as a baseline for traditional batch learning. 




\subsubsection{Results}
The \ac{GDM} architecture aims to learn distinguishable feature representations for each class, making learning class-order agnostic. In practice, however, for \ac{FER} we found the model’s performance to be sensitive to the order of learning different classes for each subject. To quantify this, we selected $6$ different class orders, starting with each of the $6$ classes used in this work. The rest of the order was randomised.

\begin{figure*}
     \centering
     \begin{subfigure}[b]{0.49\textwidth}
         \centering
         \hspace*{-0.7cm}
         \includegraphics[width=1.06\textwidth]{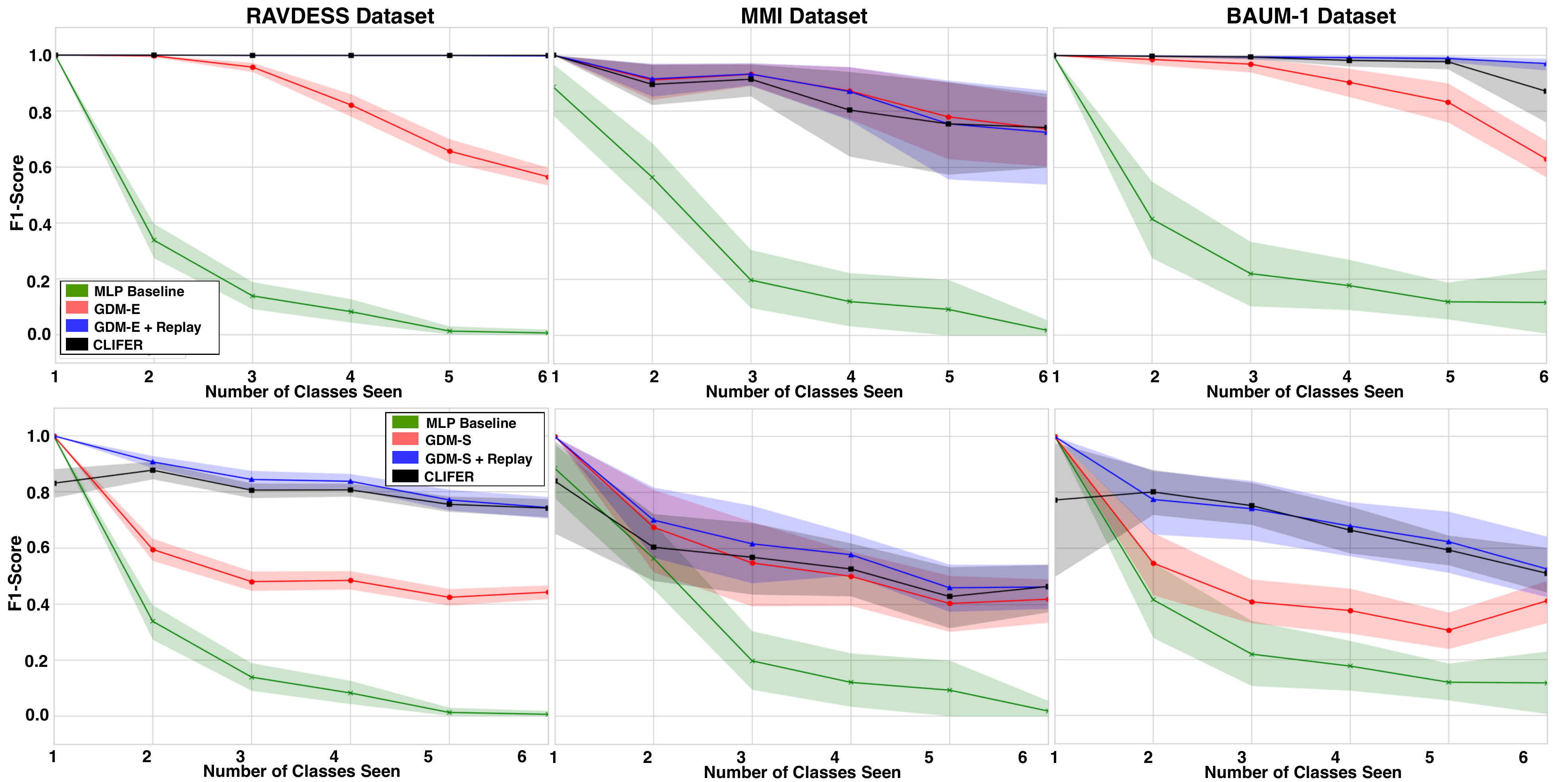}
         \caption{Experiment~1: \ac{GDM}-~E (top) and \ac{GDM}-S (bottom) performance on Remembering Seen Classes.}
         \label{fig:incremental}
     \end{subfigure}
     \begin{subfigure}[b]{0.49\textwidth}
         \centering
         \hspace*{-0.2cm}
         \includegraphics[width=1.05\textwidth]{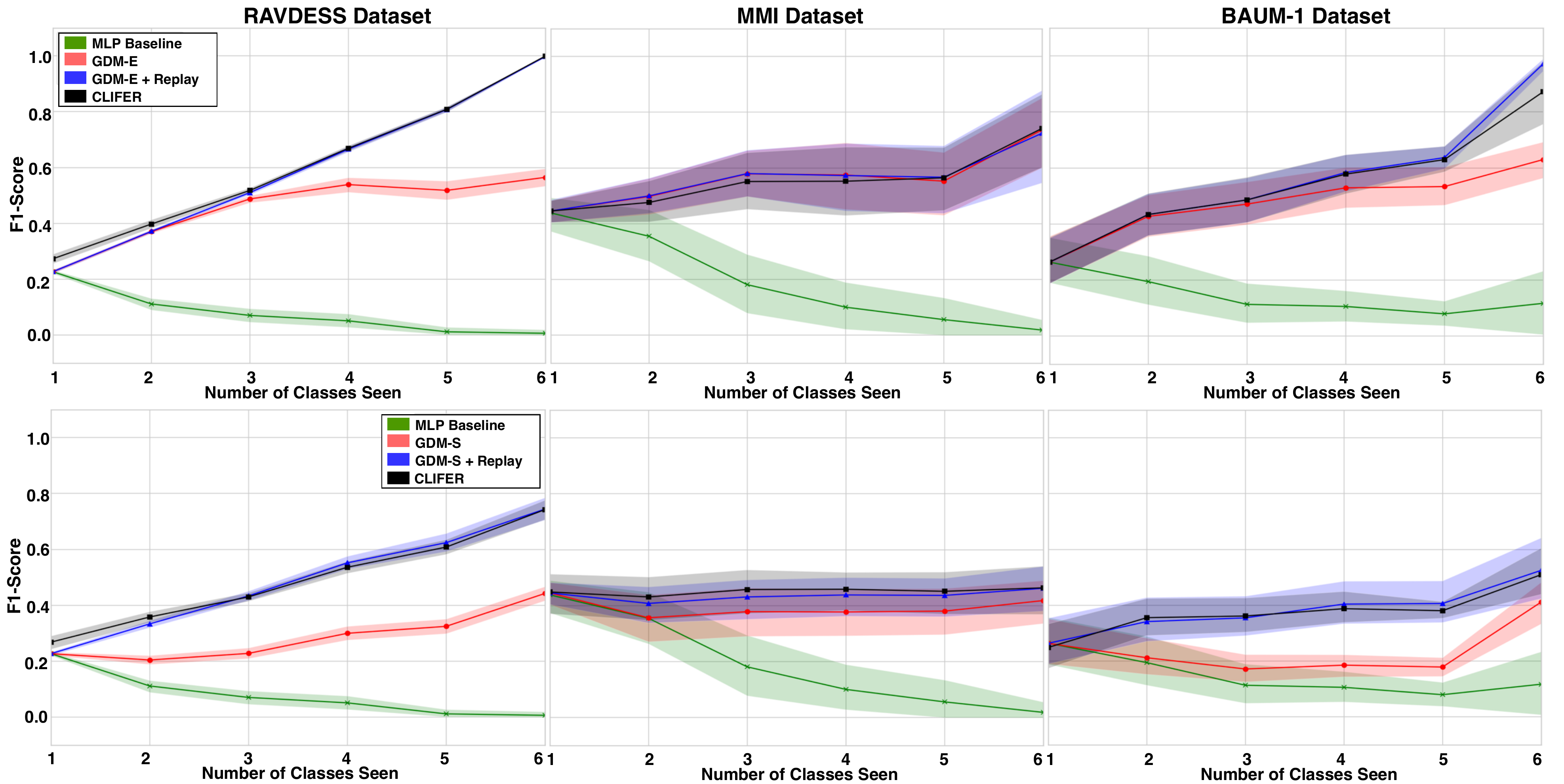}
         \caption{Experiment~2: \ac{GDM}-~E (top) and \ac{GDM}-S (bottom) performance on Generalising to New Unseen Classes.}
         \label{fig:overall}
     \end{subfigure}
     \vspace{4mm}
     \caption{F1-Scores with 95\% confidence intervals on RAVDESS, MMI and BAUM-1 datasets. Adapted from~\cite{Churamani2020CLIFER}.
     }
\end{figure*}

Kruskal-Wallis H-test results show a significant difference ($p<0.05$) in model performance for Experiment $2$ between the $6$ class orders with starting with \textit{neutral} resulting in the best performance, on average. A similar effect is seen for Experiment $1$. As the model learns how one individual expresses different emotions, the learnt feature representations overlap significantly resulting in the order impacting model performance. Other approaches in curriculum-based learning~\cite{Gui2017Curriculum}, that focus on learning facial expressions one class at a time, have also witnessed a specific order of learning (starting with high-intensity samples) enhancing model performance although they do not evaluate the models for continual learning. Starting with \textit{neutral} could be beneficial for \ac{CL}-bsed \ac{FER} models for two reasons. Firstly, \textit{neutral} represents a baseline for an individual's expressions and learning this norm enables the model to form distinct prototypes for subsequent samples that differ from this baseline. Secondly, as \textit{imagination} impacts model performance, imagined images can carry forward some of the features from the original image. Starting with \textit{neutral}, however, results in the least influence of the original image.

As a result, for our experimentation, we set the order of learning classes to start with \textit{neutral}, followed by (randomly selected) \textit{happy, surprise, anger, fear} and \textit{sadness}. The results for the RAVDESS, MMI and BAUM-1 datasets, averaged across all subjects, for the two experiments can be seen in Fig.~\ref{fig:incremental} and Fig.~\ref{fig:overall}, respectively. The \ac{GDM} model outperforms the \ac{MLP} baseline for all the $3$ datasets for both the experiments. The \ac{GDM} + Replay and the proposed CLIFER framework (that is, \ac{GDM} with \textit{imagination}) perform better than the standard \ac{GDM} model, with CLIFER, on average, performing the best across all settings resulting in high F1-scores: RAVDESS (episodic: F1$=0.98\pm0.01$, semantic: F1$=0.75 \pm 0.01$), MMI (episodic: F1$=0.75 \pm 0.07$, semantic: F1$=0.46 \pm 0.04$) and BAUM-1 (episodic: F1$=0.87 \pm 0.05$, semantic: F1$=0.51 \pm 0.04$).

These results highlight the framework's ability to adapt to an individual subject, extending its knowledge to novel classes while retaining performance on previously learnt classes. The model performance is comparable (if not better) to the state-of-the-art for RAVDESS ($79\%$~\cite{He2019Human}), MMI ($78\%$~\cite{Li2018Deep}) and BAUM-1 ($47\%$~\cite{ZhalehpourBaum2017}). Yet, it will not be correct to compare these scores directly as they do not use incremental learning for training the models and have all the data available to them apriori. Furthermore, one thing to note here is that, we select a sub-set of subjects from these datasets that provide data samples for each of the $6$ expression classes~\cite{Churamani2020CLIFER}. More experimentation is needed to further establish the performance of the CLIFER framework for all the subjects.

\section{Future Plans}
\subsection{Expanding \ac{CLIFER}}
Experimentation with \ac{CLIFER}~\cite{Churamani2020CLIFER}, as discussed above, substantiated the applicability of \ac{CL} approaches for affect perception. Yet, one thing to be noted here is that such an application of \ac{CL} is not as straightforward as other learning tasks, for example, as learning to classify objects. Important aspects such as the order of learning and context have a huge impact on learning to classify facial expressions. Furthermore, human expressions should not be viewed as isolated instances of occurrence, particularly in real-world interactions, but need to be understood as context-driven responses that evolve over a period of time in response to affective stimuli. Thus, accounting for such a temporal evolution of expression becomes crucial in recognising the affective state expressed by a user. 

Furthermore, adopting a lifelong and adaptive view on affect modelling, it is important not only recognise expressions but also model the person's long-term behaviour. Analysing how their affective behaviour evolves over time, the model can learn not just their \textit{expressions} but also estimate the \textit{mood} of an individual during an interaction as well as their long-term \textit{personality}. We are currently exploring recurrent and self-organising neural models for spatio-temporal feature learning that can enable modelling the affective state of an individual at varying temporal resolutions. Based on neuro-inspired mechanisms for affective learning, that is, the interplay between the short-term and long-term memory models in our brain that contribute towards affective association~\cite{LaBar1999Arousal,E.A.2004}, these models will fit well with the \ac{CLIFER} framework, extending it towards a \textit{multi-memory} set-up that evaluate affective expressions at different temporal resolutions.

\subsection{\ac{CLIFER} for Human-Robot Interaction}
Real-world human-robot interactions provide the best application conditions for \ac{CLIFER} as they require robots to adapt to the dynamics of each interaction, offering personalised interactions to the users. In particular, longitudinal interactions, where a user and the robot interact with each other repeatedly, over several interactions, require the agent to incrementally improve its understanding of user behaviour. In such interactions, \ac{CLIFER}, after each interaction, should be able to \textit{imagine} the user under different interaction conditions and update its learning to improve its performance for each subsequent interaction round.

To evaluate such personalised affect perception models, we will conduct a user-study with the Pepper Robot\footnote{\url{https://www.softbankrobotics.com/us/pepper}}. The user-study will involve participants repeatedly interacting with Pepper over multiple interaction sessions. Each session will be designed in a manner that it elicits a specific affective response (for example, \textit{anger} or \textit{happiness}) from the user. The task for the robot will be to learn to recognise the user's expression, personalising towards their expressions. Two conditions will be compared. In the first condition, Pepper will use a state-of-the-art \ac{FER} model while the second condition will implement the \ac{CLIFER} framework for affect perception. It is expected that the \ac{CLIFER} framework should perform better in recognising the facial expressions of the users, even after only a few interactions, and also incrementally improve its performance.

\subsection{Challenges}
One of the key challenges faced for person-specific adaptation is the lack of long-term interaction data for training the models. Most of the existing datasets, even if recording spontaneous expressions, consist of data recorded for different individuals only over a handful of interaction sessions. Also, as these interaction sessions are usually recorded all together, the data does not enable modelling affective behaviour dynamics of the individual, over time.

In the \ac{CLIFER} framework, we tackle the issue with lack of data by using the \textit{imagination} model. It enables us to generate additional data for an individual for the different expression classes. Yet, this may restrict the model's capability to handling only a few expression classes. Thus, our future work plans to extend the framework to recognising different \acp{AU}, which will enable adaptation to a wide variety of facial expressions. Additionally, we aim to consider dimensional (valence-arousal) information, moving away from categorical labels to enhance the applicability of the framework to the real-world.

\bibliographystyle{IEEEtran}
\bibliography{main.bib}

\begin{acronym}
\acro{AAE}{Adversarial Auto-Encoder}
\acro{AC}{Affective Computing}
\acro{AI}{Artificial Intelligence}
\acro{AU}{Action Unit}
\acro{BLSTM}{Bidirectional Long Short Term Memory}
\acro{BMU}{Best Matching Unit}
\acro{CAAE}{Conditional Adversarial Auto-Encoder}
\acro{CAL}{Continual Affective Learning}
\acro{cGAN}{Conditional GAN}
\acro{CHL}{Competitive Hebbian Learning}
\acro{CCC}{Concordance Correlation Coefficient}
\acro{CL}{Continual Learning}
\acro{CLIFER}{Continual Learning with Imagination for Facial Expression Recognition}
\acro{CLS}{Complementary Learning System}
\acro{CNN}{Convolutional Neural Network}

\acro{DDPG}{Deep Deterministic Policy Gradients}
\acro{DEN}{Dynamically Expanding Networks}
\acro{EWC}{Elastic Weight Consolidation}
\acro{FACS}{Facial Action Coding System}
\acro{FEL}{Fixed Expansion Layer}
\acro{FER}{Facial Expression Recognition}
\acro{GAN}{Generative Adversarial Network}
\acro{GDM}{Growing Dual Memory}
\acro{GWR}{Growing When Required}
\acro{HCI}{Human-Computer Interaction}
\acro{HHI}{Human-Human Interaction}
\acro{HRI}{Human-Robot Interaction}
\acro{LwF}{Learning Without Forgetting}
\acro{LSTM}{Long Short-Term Memory}
\acro{MFCC}{Mel-frequency Cepstral Coefficients}
\acro{ML}{Machine Learning}
\acro{MLP}{Multilayer Perceptron}
\acro{NDL}{Neurogenesis Deep Learning}
\acro{NLP}{Natural Language Processing}
\acro{OMG-Emotion}{One Minute Gradual-Emotion}
\acro{PAD}{Pleasure-Arousal-Dominance}
\acro{PFC}{Pre-Frontal Cortex}
\acro{RAVDESS}{Ryerson Audio-Visual Database of Emotional Speech and Song}
\acro{R-CNN}{Recurrent Convolutional Neural Network}
\acro{ReLU}{Rectified Linear Unit}
\acro{REM}{Rapid Eye Movement}
\acro{RL}{Reinforcement Learning}
\acro{SAR}{Socially Assistive Robotics}
\acro{SAL-DB}{Sensitive Artificial Listener Database}
\acro{SER}{Speech Emotion Recognition}
\acro{SOINN}{Self-Organizing Incremental Neural Network}
\acro{SOM}{Self-Organising Map}
\acro{SVM}{Support-Vector Machines}
\acro{SVR}{Support-Vector Machines for Regression}
\acro{TD}{Temporal Difference}
\acro{VAE}{Variational Auto-Encoder}
\acro{VCL}{Variational Continual Learning}
\acro{VI}{Variational Inference}
\end{acronym}
\end{document}